\documentclass{article}

\usepackage{arxiv}

\usepackage[utf8]{inputenc}
\usepackage[T1]{fontenc}
\usepackage{hyperref}
\usepackage{url}
\usepackage{booktabs}
\usepackage{amsfonts}
\usepackage{nicefrac}
\usepackage{microtype}
\usepackage{graphicx}
\usepackage[numbers]{natbib}
\usepackage{doi}

\usepackage{amsmath,amssymb}
\usepackage{siunitx}
\usepackage{tabularx}
\usepackage{array}
\usepackage{adjustbox}
\usepackage{tikz}
\usetikzlibrary{arrows.meta,positioning,calc,fit}

\renewcommand{\headeright}{}
\renewcommand{\undertitle}{}

\title{Multi-Head Residual-Gated DeepONet for Coherent Nonlinear Wave Dynamics}

\author{
  Zhiwei Fan\thanks{Corresponding authors: Zhiwei Fan and Daniel Coca.} \\
  School of Engineering\\
  Newcastle University\\
  NE1 7RU, United Kingdom \\
  \texttt{zhiwei.fan@ncl.ac.uk} \\
  \And
  Yiming Pan \\
  School of Physical Science and Technology\\
  ShanghaiTech University\\
  Shanghai 200031, China \\
  \texttt{yiming.pan@shanghaitech.edu.cn} \\
  \And
  Daniel Coca\footnotemark[1] \\
  School of Engineering\\
  Newcastle University\\
  NE1 7RU, United Kingdom \\
  \texttt{daniel.coca@ncl.ac.uk} \\
}

\date{}

\renewcommand{\shorttitle}{MH-RG DeepONet for Coherent Nonlinear Wave Dynamics}

\hypersetup{
pdftitle={Multi-Head Residual-Gated DeepONet for Coherent Nonlinear Wave Dynamics},
pdfauthor={Zhiwei Fan, Yiming Pan, Daniel Coca},
pdfkeywords={DeepONet, operator learning, coherent nonlinear dynamics, physics-informed machine learning},
}

\begin{document}
\maketitle

\begin{abstract}
Coherent nonlinear wave dynamics are often strongly shaped by a compact set of physically meaningful descriptors of the initial state. Traditional neural operators typically treat the input-output mapping as a largely black-box high-dimensional regression problem, without explicitly exploiting this structured physical context. Common feature-integration strategies usually rely on direct concatenation or FiLM-style affine modulation in hidden latent spaces. Here we introduce an explicit two-pathway conditioning strategy in which the full initial field is encoded by the DeepONet state pathway, while compact, sample-specific physical descriptors act through a parallel modulation pathway.

Based on this idea, we develop a Multi-Head Residual-Gated DeepONet (MH-RG), which combines a pre-branch residual modulator, a branch residual gate, and a trunk residual gate with a low-rank multi-head mechanism to capture multiple complementary conditioned response patterns without prohibitive parameter growth. We evaluate the framework on representative benchmarks including highly nonlinear conservative wave dynamics and dissipative trapped dynamics and further perform detailed mechanistic analyses of the learned multi-head gating behavior. A complete component ablation identifies the pre-branch module as the dominant individual conditioning pathway, with the branch- and trunk-side residual gates providing additional latent refinement. Controlled descriptor experiments show that the physically defined context outperforms learned and random contexts, while single-head, shared-gate multi-readout, and head-specific controls establish an additional contribution from non-identical, head-dependent conditioning. 
Compared with conditioned DeepONet baselines and the matched
attention-based operator, MH-RG achieves consistently lower
error while better preserving phase-sensitive field structure and the
fidelity of physically relevant dynamical quantities.
\end{abstract}


\section{Introduction}

Coherent nonlinear wave dynamics arise across a broad range of physical systems, including nonlinear optics, ultracold atomic gases, water-wave soliton media, nonlinear field theories governed by the sine-Gordon family of equations and so on. Representative examples include optical solitons and Kerr frequency combs in nonlinear resonators~\cite{Agrawal2019,Hasegawa1995,DelHaye2007,Kippenberg2018,Fan2022,NA2023}, coherent matter-wave excitations, soliton trains, and breathing states in Bose-Einstein condensates~\cite{Strecker2002,Khaykovich2002,Nguyen2017}. Although these systems differ in microscopic origin and governing equations, they share a common feature: their evolution is often strongly shaped by a relatively compact set of physically meaningful descriptors of the initial state, such as total intensity or energy, localization width, center position, momentum, or spectral spread. For such systems, the challenge is not merely to fit a generic high-dimensional input-output map, but to learn an operator that remains sensitive to the physically structured mechanisms governing coherent propagation, interaction, deformation, trapping, and dissipation.

Learning fast and accurate surrogates for nonlinear partial differential equations (PDEs) is a central objective in scientific machine learning, with applications spanning nonlinear optics, quantum many-body systems, nonlinear wave equations, and fluid-like dynamics~\cite{Karniadakis2021Review,Kovachki2023JMLR}. Physics-Informed Neural Networks (PINNs)~\cite{Raissi2019} showed that governing equations can be incorporated directly into optimization, but training may become challenging in strongly nonlinear or stiff regimes, where optimization pathologies and spectral bias can hinder fidelity and stability~\cite{Wang2021GradientPathologies,Wang2022AdaptivePINN,Ji2021Stiff}. Neural operators such as DeepONet~\cite{lu2021deeponet}, the Fourier Neural Operator (FNO)~\cite{Li2021FNO}, and Physics-Informed Neural Operators (PINO)~\cite{Li2022PINO} provide an alternative paradigm by learning function-to-function mappings directly from data, and attention-based variants further broaden the range of accessible dynamics~\cite{Hao2023ICML,Li2023OFormer}. More recent studies have also explored invariant or physics-aware operator constructions~\cite{Liu2023INO,Zhang2024PIANO}, long-time integration of wave dynamics~\cite{Lei2024LongTime}, fair benchmarking of neural operator families~\cite{lu2021deeponet_fair}, and reconstruction strategies for PDEs with nonsmooth or discontinuous solutions~\cite{Samuel2023}. These challenges are especially pronounced for phase-sensitive complex fields: the real and imaginary channels jointly encode phase, nonlinear collisions generate sharp local phase variation and spectral broadening, and a pointwise field loss does not directly constrain mass, Hamiltonian, momentum, or relative phase. 

This is particularly relevant for coherent nonlinear wave dynamics. In many such systems, the long-time or strongly nonlinear evolution is not governed by arbitrary fine-scale fluctuations alone, but remains organized around a small number of collective descriptors extracted from the initial state. In nonlinear fiber optics, for example, peak amplitude, pulse width, carrier momentum, and spectral bandwidth can strongly influence soliton interactions, dispersive radiation, and comb formation. In trapped Bose-Einstein condensates, center-of-mass displacement, total mass, and width-like quantities help organize breathing, trapping, and dissipative relaxation.  This suggests that, for an important class of physically structured problems, operator learning should benefit from explicitly combining two complementary ingredients: a high-dimensional representation of the evolving state, and a low-dimensional physically interpretable context that modulates how this state representation is propagated. This inspiration leads to architectural separation: the full field remains the state input to the operator, whereas a compact descriptor vector provides an auxiliary, physically interpretable conditioning signal. 

Motivated by this viewpoint, we introduce the Multi-Head Residual-Gated DeepONet (MH-RG DeepONet) for coherent nonlinear wave dynamics. The architecture retains the standard DeepONet branch-trunk factorization for state encoding and coordinate encoding, but augments it with three modulation components driven by the same compact physical descriptor vector extracted from the initial condition: a pre-branch modulator acting directly on the sensor input, a branch residual gate acting on the latent state embedding, and a trunk residual gate acting on the latent coordinate embedding. The residual gate can be viewed as a constrained multiplicative FiLM-like transformation with zero additive shift and an identity-centred, bounded scale. Its architectural distinction from the standard branch-hidden FiLM baseline~\cite{Perez2018FiLM} lies primarily in the placement of conditioning: an input-side pre-branch modulator is combined with separate branch- and trunk-side latent gates. In our formulation, the conditioning signal acts as a residual multiplicative modulator that preserves a strong identity path and intervenes structurally at the input-side and latent branch/trunk interaction levels, rather than serving as a generic affine perturbation of hidden activations.

Our main model further extends this residual-gated design through a low-rank multi-head mechanism. The rationale is twofold. First, a single conditioned interaction pathway may be too rigid to represent the full space--time response, whereas a multi-head decomposition allows several conditioned response components to coexist and complement one another. We do not assume a one-to-one correspondence between individual heads and uniquely identifiable physical processes; rather, the heads are interpreted as non-identical computational response channels. Second, a naive multi-head extension with fully independent dense gates would incur substantial parameter growth and blur the distinction between structured conditioning and brute-force overparameterization. We therefore introduce shared low-rank gate manifolds, so that the different heads explore distinct conditioned directions within a constrained latent subspace rather than expanding into fully unconstrained dense modulation.

We evaluate the proposed framework on two representative case studies spanning distinct classes of coherent nonlinear dynamics: a conservative nonlinear Schr\"odinger benchmark motivated by nonlinear optics, and a dissipative breathing-state problem in a trapped Bose-Einstein condensate. These two examples are chosen to probe qualitatively different forms of structured wave evolution, including conservative versus dissipative dynamics, localized collision-dominated interactions versus trapped collective breathing, and strongly phase-sensitive complex-field propagation in both settings. Across these cases, the results clarify how different forms of physics-guided conditioning introduce different inductive biases into operator learning. In particular, the experiments show that the complete pre-branch/branch/trunk architecture outperforms direct concatenation and the standard branch-hidden FiLM baseline. The complete factorial ablation demonstrates that the bounded residual branch and trunk pathways make individually measurable contributions, while another principal additional gain is obtained from the low-rank head-specific decomposition. 

The contributions of this work are below. First, we formulate coherent nonlinear dynamics as a natural setting in which compact physical descriptors of the initial state can be used as structured context for operator learning.   Second, we introduce an input-side pre-branch modulator together with bounded, identity-centred branch and trunk residual gates; complete factorial ablation identifies the pre-branch pathway as the dominant individual module and the latent gates as additional refinements. Third, we develop a shared-low-rank multi-head decomposition and isolate its contribution using single-head and shared-gate multi-readout controls and dataset-level statistics. 

\begin{figure*}[h!]
\centering
\includegraphics[width=0.95\textwidth]{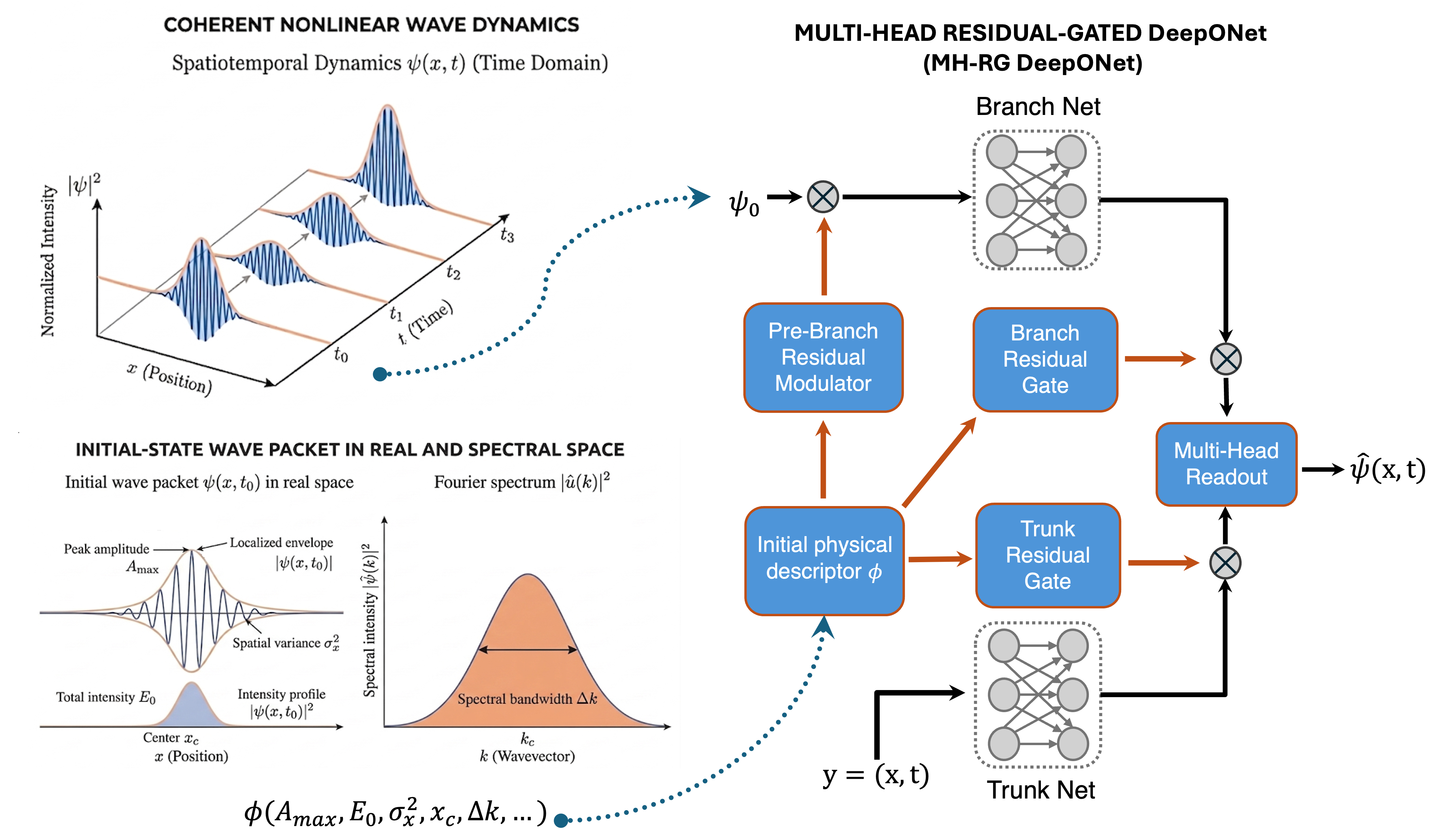}
\caption{\textbf{Physical motivation and architecture of the MH-RG DeepONet framework.}
\textbf{Left:} For coherent nonlinear dynamics, the future evolution of the field \(\psi(x,t)\) is often influenced not only by the full initial state \(\psi(x,t_0)\), but also by a compact set of physically meaningful descriptors extracted from that initial condition. Examples include peak amplitude, total intensity, center position, spatial variance, spectral bandwidth and so on, which summarize the structured content of the initial field in real and spectral space.
\textbf{Right:} The proposed MH-RG DeepONet combines these two sources of information. The full initial state is processed by the Branch network, while the coordinate query \((x,t)\) is processed by the Trunk network. In parallel, the descriptor vector \(\phi\) modulates the operator through three residual conditioning pathways: a pre-branch residual modulator, a branch residual gate, and a trunk residual gate. The resulting conditioned branch and trunk embeddings are fused and passed to a multi-head readout to produce the prediction \(\hat \psi(x,t)\).}
\label{fig:architecture}
\end{figure*}

 \section{Methodology}

We build our model family on the standard DeepONet backbone~\cite{lu2021deeponet}. 
Let \(\mathcal{G}\) denote the nonlinear solution operator that maps an initial condition \(\psi_0\) to the complex-valued field evaluated at a queried space-time coordinate:
\[
\mathcal{G}: \psi_0 \mapsto \psi(x,t).
\]
Let \(\{x_i\}_{i=1}^{N_x}\) denote the spatial grid. 
The complex-valued initial condition \(\psi_0(x_i)\) is represented by concatenating its real and imaginary parts into a sensor vector
\[
s(\psi_0)\in\mathbb{R}^{2N_x}.
\]
The branch network \(B\) maps this sensor representation to a latent state embedding
\[
\mathbf{b}=B(s)\in\mathbb{R}^{p},
\]
where \(p\) is the latent width. 
For each query point \((x_j,t_j)\), the trunk network \(T\) produces a coordinate embedding
\[
\mathbf{t}_j=T(x_j,t_j)\in\mathbb{R}^{p}.
\]
For a \(q\)-channel output (here \(q=2\), corresponding to the real and imaginary parts), the vanilla DeepONet prediction is
\begin{equation}
    \hat{\mathbf{\psi}}(x_j,t_j)
    \;=\;
    W^\top\!\big(\mathbf{b}\odot\mathbf{t}_j\big) + \mathbf{c},
    \qquad
    W\in\mathbb{R}^{p\times q},\; \mathbf{c}\in\mathbb{R}^q,
    \label{eq:vanilla-deeponet}
\end{equation}
where \(\odot\) denotes the Hadamard product.

\subsection{Initial-state descriptors}

In addition to the full initial condition, we compute from each sample a compact feature vector
\[
\phi(\psi_0)\in\mathbb{R}^{N_f},
\]
which serves as a low-dimensional summary of the initial state and $N_f$ denotes the number of descriptors. 
For coherent wave dynamics, such as a typical propagating wavepacket in Fig.~\ref{fig:architecture}, they provide a compact conditioning signal that summarizes leading global properties of the initial waveform, including overall intensity, localization, translation tendency, and spectral spread.

These quantities are not introduced as arbitrary engineered statistics. Norm or mass, peak amplitude, centre and variance, momentum, and spectral width are standard experimentally accessible observables or routinely evaluated collective diagnostics in nonlinear optics and matter-wave physics~\cite{Agrawal2019,Sulem1999Springer,Choi1998PRA,Reeves2012PRA}. They organize physically distinct aspects of coherent dynamics, including global intensity scale, localization and compression, translational motion, phase tilt, and occupied wave-number range. The full field remains the state input to the branch network; the descriptor vector supplies a physically privileged summary rather than replacing the state.

The concrete descriptor list is PDE dependent rather than universal. We select candidate quantities from four physically motivated groups: (i) conserved quantities or balance-law variables, such as mass, norm, or energy; (ii) collective coordinates associated with symmetries and transport, such as centre of mass and momentum; (iii) localization and scale measures, such as peak amplitude and spatial variance; and (iv) spectral statistics, such as spectral centroid and bandwidth. A descriptor is retained only when it can be computed from the available initial observation, is numerically stable and low dimensional, and is relevant to the dominant dynamical regimes. The learned-context control shows that the architecture can operate without predefined observables, whereas the lower error of the physical context quantifies the data-efficient inductive bias supplied by domain knowledge. The descriptor vector is therefore not assumed to be a mathematically sufficient state representation.

\subsection{Conditioned DeepONet and attention-based baselines}

To separate the effect of merely providing auxiliary features from the effect of using them in a structured manner, we compare several conditioned DeepONet variants studied in the nonlinear Schr\"odinger model, summarized in Table~\ref{tab:arch_spec}.

The Concat baseline injects the full descriptor vector directly into the branch input:
\[
[s(\psi_0),\phi(\psi_0)]\in\mathbb{R}^{2N_x+N_f},
\]
while the trunk network still takes only the query coordinates \((x,t)\). 
This provides a simple black-box feature-injection baseline without explicit modulation.

The FiLM baseline applies canonical feature-wise affine modulation~\cite{Perez2018FiLM, GEK2026} to the hidden layers of the branch network. 
At each branch hidden layer, a conditioning network maps \(\phi(\psi_0)\) to feature-wise scale and bias coefficients, so that the hidden activations are modulated as
\[
h'=\gamma(\phi)\odot h+\beta(\phi).
\]
This provides a strong standard conditioned baseline in which the initial-state descriptors act through affine hidden-feature modulation, while the trunk pathway remains unchanged.

As an external attention-based neural-operator baseline, we additionally implement a parameter-matched  OFormer~\cite{Li2023OFormer}. The adaptation follows the one-dimensional OFormer configuration and contains four Galerkin self-attention encoder blocks, an eight-head Fourier cross-attention decoder with relative rotary coordinate encoding, Gaussian Fourier query features, and three non-shared residual propagator MLPs. The input token at sensor location \(x_i\) is \([\mathrm{Re}\,\psi_0(x_i),\mathrm{Im}\,\psi_0(x_i),x_i]\), and the model predicts the two output channels at arbitrary queries \((x,t)\).

\subsection{Residual-gated DeepONet (RG)}

Our residual-gated baseline extends DeepONet by introducing three conditioning modules driven by the same feature vector \(\phi\): a pre-branch residual modulator, a branch residual gate, and a trunk residual gate.

Before branch encoding, the sensor representation is modulated channel-wise as
\begin{equation}
    \tilde{s}
    \;=\;
    s \odot \Big(
        \mathbf{1}_{2N_x}
        + \alpha_{\mathrm{pre}} \tanh\!\big(h_{\mathrm{pre}}(\phi)\big)
    \Big),
    \label{eq:prebranch-new}
\end{equation}
where \(h_{\mathrm{pre}}:\mathbb{R}^{N_f}\to\mathbb{R}^{2N_x}\) is a small MLP and \(\alpha_{\mathrm{pre}}>0\) controls the modulation strength. 
This module acts directly on the input sensor before nonlinear branch encoding, and may be viewed as a residual input-side modulator that reweights the sensor channels according to the global initial-state descriptors.

After branch and trunk encoding, we apply feature-conditioned residual gates to the branch and trunk embeddings:
\begin{equation}
\begin{split}
    \gamma_B(\phi)
    \;=\;
    \mathbf{1}_p + \alpha_B \tanh\!\big(f_B(\phi)\big)\in\mathbb{R}^p, \\
    \gamma_T(\phi)
    \;=\;
    \mathbf{1}_p + \alpha_T \tanh\!\big(f_T(\phi)\big)\in\mathbb{R}^p,
\end{split}
\label{eq:single-gates-new}
\end{equation}
where \(f_B:\mathbb{R}^{N_f}\to\mathbb{R}^{p}\) and \(f_T:\mathbb{R}^{N_f}\to\mathbb{R}^{p}\) are small MLPs. 
The branch and trunk embeddings are then modulated as
\[
\tilde{\mathbf{b}}
    \;=\;
    B(\tilde{s})\odot\gamma_B(\phi),
    \qquad
\tilde{\mathbf{t}}_j
    \;=\;
    T(x_j,t_j)\odot\gamma_T(\phi).
\]
The final prediction becomes
\begin{equation}
    \hat{\mathbf{\psi}}_{\mathrm{RG}}(x_j,t_j)
    \;=\;
    W^\top\!\big(\tilde{\mathbf{b}}\odot\tilde{\mathbf{t}}_j\big) + \mathbf{c}.
    \label{eq:single-rg-readout}
\end{equation}

Because the gate is bounded and centered around the identity, it preserves a clear identity path and makes the conditioning explicitly modulatory, instead of replacing the latent representation outright.
More precisely, the residual gate is a constrained special case of feature-wise modulation. This design allows the combination of an input-side pre-branch modulation pathway with separate branch- and trunk-side bounded corrections. The residual form provides a simple auxiliary route that can approach the identity when descriptor-dependent correction is unnecessary and restricts the scale to \([1-\alpha,1+\alpha]\). In our framework, the physical descriptors are treated as a compact auxiliary channel that corrects the learned operator representation without redefining it.

\subsection{Multi-head residual-gated DeepONet (MH-RG)}

Our main model extends the residual-gated design by introducing a multi-head decomposition at the latent-gating and readout levels. 
As in Eq.~\eqref{eq:prebranch-new}, the sensor input is first modulated by the pre-branch residual modulator,
 after which the shared branch and trunk embeddings are computed as
\[
\mathbf{b}=B(\tilde{s}),\qquad
\mathbf{t}_j=T(x_j,t_j).
\]

Let \(R\) denote the number of heads and \(r_g\ll p\) a low-rank gate dimension. 
To increase expressivity without introducing fully independent dense gates for every head, we use two shared upsampling matrices
\[
U_B\in\mathbb{R}^{p\times r_g},
\qquad
U_T\in\mathbb{R}^{p\times r_g}.
\]
For each head \(r\in\{1,\dots,R\}\), we compute feature-conditioned branch and trunk gate pre-activations as
\begin{equation}
\begin{split}
f_B^{(r)}(\phi)
  \;=\;
  U_B\,h_B^{(r)}(\phi)\in\mathbb{R}^{p}, \\
f_T^{(r)}(\phi)
  \;=\;
  U_T\,h_T^{(r)}(\phi)\in\mathbb{R}^{p},
\end{split}
\label{eq:mhog-preact-new}
\end{equation}
where
\[
h_B^{(r)}:\mathbb{R}^{N_f}\to\mathbb{R}^{r_g},
\qquad
h_T^{(r)}:\mathbb{R}^{N_f}\to\mathbb{R}^{r_g}
\]
are small head-specific MLPs. 
The corresponding residual gates are
\begin{equation}
\begin{split}
\gamma_B^{(r)}(\phi)
  \;=\;
  \mathbf{1}_p + \alpha_B\,\tanh\!\big(f_B^{(r)}(\phi)\big), \\
\gamma_T^{(r)}(\phi)
  \;=\;
  \mathbf{1}_p + \alpha_T\,\tanh\!\big(f_T^{(r)}(\phi)\big).
\end{split}
\label{eq:mhog-gates-new}
\end{equation}
Each head produces a head-specific prediction

\begin{equation}
    \hat{\mathbf{\psi}}^{(r)}(x_j,t_j)
    \;=\;
    W_r^\top
    \Big[
        \big(\mathbf{b}\odot\gamma_B^{(r)}(\phi)\big)
        \odot
        \big(\mathbf{t}_j\odot\gamma_T^{(r)}(\phi)\big)
    \Big] +\mathbf{c}_r,
    \label{eq:mhog-head-new}
\end{equation}
where \(W_r\in\mathbb{R}^{p\times q}\) and
\(\mathbf{c}_r\in\mathbb{R}^{q}\) are the readout matrix and bias of head \(r\).
The final prediction is the sum of all head contributions:
\begin{equation}
    \hat{\mathbf{\psi}}(x_j,t_j)
    \;=\;
    \sum_{r=1}^{R}\hat{\mathbf{\psi}}^{(r)}(x_j,t_j).
    \label{eq:mhog-sum-new}
\end{equation}

This yields a multi-head residual-gated DeepONet, in which multiple conditioned response components are represented in parallel and then superposed.
The low-rank construction may be interpreted as a shared conditional gate dictionary. Writing
\(U_B=[u_{B,1},\ldots,u_{B,r_g}]\), the branch-side gate preactivation of head \(r\) is
\[
f_B^{(r)}(\phi)=\sum_{\ell=1}^{r_g}h_{B,\ell}^{(r)}(\phi)u_{B,\ell},
\]
and the trunk side has the same form. Thus, the heads do not learn \(R\) unrestricted \(p\)-dimensional gate generators. They learn head-specific nonlinear coefficient maps over shared \(r_g\)-dimensional branch and trunk bases. This creates a controlled intermediate regime between fully shared conditioning and fully independent dense experts.
Ignoring biases, a dense head-specific branch/trunk gate construction with gate hidden width \(h_g\) has leading complexity
\[
N_{\mathrm{dense}}=\mathcal{O}\!\left(2R(N_fh_g+h_gp)\right),
\]
whereas the shared-low-rank construction has
\[
N_{\mathrm{low-rank}}=\mathcal{O}\!\left(2pr_g+2R(N_fh_g+h_gr_g)\right).
\]
For \(p=256\), \(h_g=32\), and \(r_g=8\), the final head-specific mapping is reduced from \(32\times256\) to \(32\times8\) weights per side and head, while the \(8\times256\) upsampler is paid only once. This is a factor of \(p/r_g=32\) compression in the per-head output map and explains why the head count can be increased with only modest growth of the full model.

The final operator is an additive expansion over descriptor-conditioned latent interactions.
This form permits several nonlinear coefficient maps and readouts to act on the same shared branch--trunk representation, but it does not mathematically force the heads to be complementary. We therefore use three controlled variants: a single-head low-rank RG model; a shared-gate multi-readout control in which all readouts use identical branch and trunk gates; and a head-specific multi-readout control using the proposed gate topology. Below, we show that the additional improvement of the head-specific multi-readout control over this shared-gate control isolates the contribution of distinct, head-dependent descriptor-to-gate mappings.
Empirically, this provides a more flexible conditioned operator approximation while avoiding the parameter growth of fully independent dense head-specific gates.

\subsection{Training protocol and normalization}

All models are trained under the same normalization protocol. 
The real and imaginary parts of the input initial conditions are normalized channel-wise using training-set statistics only. 
Specifically, if \(\psi_0=\psi_0^{(R)}+i\,\psi_0^{(I)}\), then the sensor input is formed from
\[
\frac{\psi_0^{(R)}-\mu_R^{\mathrm{in}}}{\sigma_R^{\mathrm{in}}},
\qquad
\frac{\psi_0^{(I)}-\mu_I^{\mathrm{in}}}{\sigma_I^{\mathrm{in}}},
\]
where \(\mu^{\mathrm{in}}\) and \(\sigma^{\mathrm{in}}\) are computed over the training set. 
The same principle is used for the output field: the real and imaginary parts of the target solution are normalized channel-wise using training-set output statistics. 
The feature vector \(\phi\) is likewise standardized using the training-set mean and standard deviation of each feature dimension.

During training, each sample is paired with randomly subsampled query points \((x_j,t_j)\), and the model predicts the normalized real and imaginary parts at those queried coordinates. 
All models are optimized with Adam using the same learning rate and training budget. 
The training loss is the mean-squared error in the normalized output space:
\[
\mathcal{L}_{\mathrm{train}}
=
\frac{1}{N}
\sum_{j=1}^{N}
\left\|
\hat{\mathbf{\psi}}_{\mathrm{norm}}(x_j,t_j)
-
\mathbf{\psi}_{\mathrm{norm}}(x_j,t_j)
\right\|_2^2.
\]
Gradient clipping is applied during optimization for stability.

At inference time, the model outputs are de-normalized back to the original physical scale using the training-set output statistics. 
All reported full-field errors in the main text and tables are therefore computed in raw physical units, not in normalized space. 
Specifically, for the complex-valued prediction \(\hat{\psi}\) and ground truth \(\psi\), we report the full-field MSE
\[
\mathrm{MSE}
=
\frac{1}{N_tN_x}
\sum_{i,j}
\left[
\big(\mathrm{Re}\,\hat{\psi}_{ij}-\mathrm{Re}\,\psi_{ij}\big)^2
+
\big(\mathrm{Im}\,\hat{\psi}_{ij}-\mathrm{Im}\,\psi_{ij}\big)^2
\right].
\]
Unless otherwise stated, all quantitative results are aggregated over three independent random seeds and reported as mean \(\pm\) standard deviation across seed means.

For a fair comparison, the conditioned variants are trained using the same dataset, input normalization, query sampling procedure, optimizer, evaluation protocol and random seeds. This consistent setup minimizes performance differences arising from training, normalization, or evaluation choices and allows the comparison to focus on the model architectures themselves.

% ===================== Model Clarification (updated final version) =====================
\begin{table*}[t]
\centering
\caption{Network architecture specification (layer widths) for the NLSE study. Each tuple $(d_0,\dots,d_L)$ denotes a feed-forward MLP with Linear layers $d_0\!\to\!\cdots\!\to\!d_L$ and GELU activations between hidden layers. Here $2N_x=256$, output dimension $q=2$, and latent width $p=256$ for all DeepONet variants. The selected conditioning descriptors are $\phi=[E_0,A_{\max},x_c,\sigma_x^2,P_0,\Delta k]\in\mathbb{R}^6$.}
\label{tab:arch_spec}

\setlength{\tabcolsep}{8pt}
\renewcommand{\arraystretch}{1.10}

% ===================== TOP: backbone =====================
\begin{tabular}{l c c c}
\toprule
\textbf{Model} & \textbf{Branch $B(\cdot)$} & \textbf{Trunk $T(\cdot)$} & \textbf{$p$} \\
\midrule
\addlinespace[2pt]
Vanilla            & $(256,512,512,512,256)$ & $(2,512,512,512,256)$ & $256$ \\
Concat             & $(262,512,512,512,256)$ & $(2,512,512,512,256)$ & $256$ \\
FiLM               & $(256,512,512,512,256)$ & $(2,512,512,512,256)$ & $256$ \\
RG                 & $(256,512,512,512,256)$ & $(2,512,512,512,256)$ & $256$ \\
MH-RG ($R$ heads)   & $(256,512,512,512,256)$ & $(2,512,512,512,256)$ & $256$ \\
\bottomrule
\end{tabular}

\vspace{8pt}
\noindent\rule{\textwidth}{1.2pt}
\vspace{8pt}

% ===================== BOTTOM: conditioning =====================
\begin{tabular}{l p{13cm}}
\toprule
\textbf{Model} & \textbf{Conditioning / gating modules} \\
\midrule

Vanilla &
No explicit conditioning. The branch network encodes the normalized sensor input $s(\psi)\in\mathbb{R}^{256}$, the trunk network encodes the query coordinates $(x,t)\in\mathbb{R}^2$, and the prediction is obtained through the standard DeepONet multiplicative fusion followed by a linear readout. \\

Concat &
No explicit gating. The full six-dimensional initial-state descriptor vector is concatenated directly to the branch input:
Branch input $[s(\psi_0),\phi]\in\mathbb{R}^{256+6}$, while the trunk input remains $[x,t]\in\mathbb{R}^{2}$.
Thus, Concat serves as a black-box feature-injection baseline without structured modulation. \\

FiLM &
Canonical FiLM baseline applied to the branch hidden layers.
At each branch hidden layer, a conditioning network maps the full descriptor vector $\phi\in\mathbb{R}^{6}$ to feature-wise affine FiLM coefficients,
\[
\mathrm{FiLM}(h\mid \phi)=\gamma(\phi)\odot h+\beta(\phi),
\]
where each layer uses a generator of size $(6,128,2\times 512)$.
Thus, FiLM modulation is applied at every hidden layer of the branch MLP, while the trunk network remains unchanged.~\cite{GEK2026} \\

RG &
Residual-gated DeepONet with a pre-branch residual modulator and latent residual gates, all conditioned on the same six-dimensional descriptor vector $\phi$.
A pre-branch residual modulator acts on the raw sensor input:
$g_{\mathrm{pre}}(\phi):(6,16,256)$ with $\alpha_{\mathrm{pre}}=1.0$.
Residual latent gates are then applied separately on the branch and trunk latents:
$g_B(\phi):(6,256,256)$ and $g_T(\phi):(6,256,256)$,
with gate amplitudes $\alpha_B=\alpha_T=0.5$. \\

MH-RG ($R$ heads) &
Multi-head residual-gated DeepONet with the same pre-branch residual modulator as RG:
$g_{\mathrm{pre}}(\phi):(6,16,256)$ with $\alpha_{\mathrm{pre}}=1.0$.
The latent gates are parameterized in low-rank multi-head form with $r_g=8$ and gate hidden width $h_g=32$:
shared upsamplers $U_B,U_T:(8\!\to\!256)$;
per-head generators $h_B^{(r)}(\phi):(6,32,8)$ and $h_T^{(r)}(\phi):(6,32,8)$;
and per-head readouts $W_r:(256\!\to\!2)$, whose outputs are summed as $\sum_{r=1}^{R}$.
The branch-side and trunk-side gate amplitudes are fixed to $\alpha_B=\alpha_T=0.5$. \\
\bottomrule
\end{tabular}
\end{table*}

\section{Experiments}\label{sec:experiments}
\subsection{Strong nonlinear interaction benchmark: focusing NLSE}

We first consider the one-dimensional focusing cubic nonlinear Schr\"odinger equation (NLSE)~\cite{Zakharov1972JETP,Sulem1999Springer}
\begin{equation}
    i\,\partial_t \psi + \tfrac{1}{2}\partial_{xx}\psi + |\psi|^2\psi = 0,
\end{equation}
on $(x,t)\in[-10,10]\times[0,2]$. Reference solutions are generated by a split-step Fourier solver \cite{Weideman1986SIAM} on a uniform grid with $(N_x,N_t)=(128,201)$. The training and test sets contain $1000$ and $200$ trajectories, respectively. The initial condition is a superposition of two modulated Gaussian wave packets,
\begin{equation}
    \psi_0(x)=\sum_{j=1}^2 A_j
    \exp\!\Big(-\frac{(x-x_j)^2}{2w_j^2}\Big)e^{ik_jx},
\end{equation}
The amplitudes are sampled as \(A_j\in[1.0,1.5]\), the widths as \(w_j\in[0.8,1.3]\), the packet centres as \(x_1\in[-5,-3]\) and \(x_2\in[3,5]\), and the carrier wavenumbers as \(k_1\in[2,4]\) and \(k_2\in[-4,-2]\). These ranges generate counter-propagating packets that undergo a fast, strongly nonlinear collision within the simulated time window. 

This benchmark contains several distinct stages within a single rollout: incoming propagation, a localized high-intensity interaction region, and an outgoing post-collision regime. Therefore, a strong surrogate should preserve the complex-field structure through the interaction, since local errors near the collision region typically reappear later as phase distortion, amplitude mismatch, or drift in derived physical quantities. This makes the NLSE collision problem a suitable test for structured conditioning by compact initial-state descriptors.

In the one-dimensional NLSE setting considered here, we use
\[
\phi(\psi_0)=
\big[E_0,\;A_{\max},\;x_c,\;\sigma_x^2,\;P_0,\;\Delta k\big]^\top.
\]
These six descriptors are defined as follows:
\begin{itemize}
    \item \(E_0=\int |\psi(x)|^2\,dx\): total intensity (or mass);
    \item \(A_{\max}=\max_x |\psi(x)|\): peak amplitude;
    \item \(x_c = E_0^{-1}\int x\,|\psi(x)|^2\,dx\): center of intensity;
    \item \(\sigma_x^2 = E_0^{-1}\int (x-x_c)^2 |\psi(x)|^2\,dx\): spatial variance;
    \item \(P_0=\mathrm{Im}\int \psi^*\,\partial_x \psi\,dx\): linear momentum;
    \item \(\Delta k\): spectral bandwidth, computed from the second moment of the Fourier power spectrum.
\end{itemize}

For this benchmark we extract six descriptors from the initial state,
\[
\phi=[E_0, A_{\max}, x_c, \sigma_x^2, P_0, \Delta k],
\]
corresponding to total intensity, peak amplitude, center of intensity, spatial variance, momentum, and spectral bandwidth. In the present all-feature setting, the same descriptor vector is supplied to all conditioning pathways of RG and MH-RG, namely the pre-branch residual modulator, the branch residual gate, and the trunk residual gate. By contrast, Concat appends the same six descriptors directly to the branch input, while FiLM applies canonical affine modulation to the hidden layers of the branch network.

\begin{figure*}[h]
\centering
\includegraphics[width=0.95\textwidth]{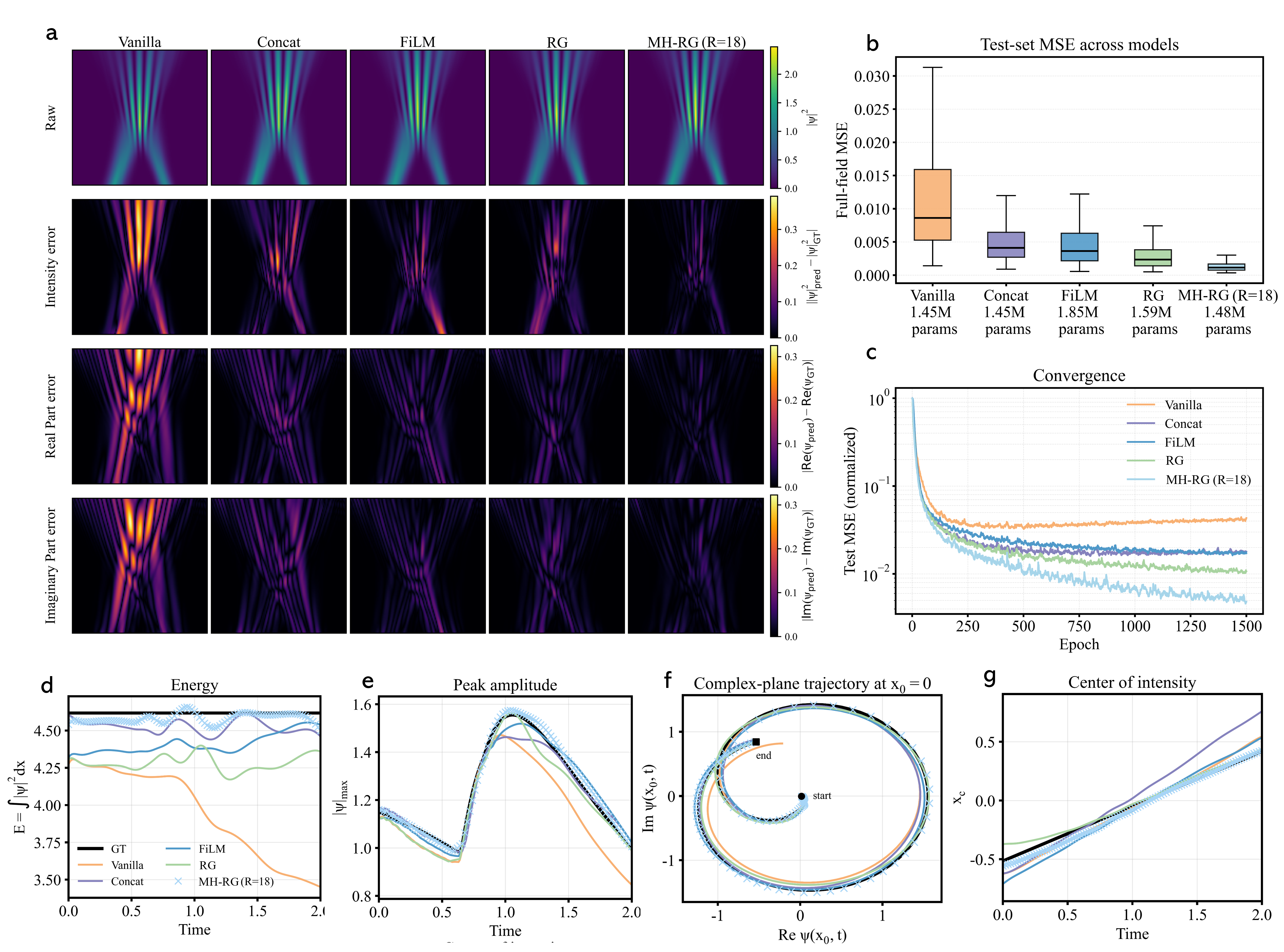}
\caption{\textbf{NLSE benchmark: field-level, optimization, and physics-level comparisons.}
(a) Representative full-field predictions and absolute error maps for Vanilla, Concat, FiLM, RG, and MH-RG ($R=18$).
Rows show the predicted intensity $|\hat{\psi}|^2$, the intensity error, the real-part error, and the imaginary-part error.
(b) Distribution of per-sample full-field MSE on the test set.
(c) Mean convergence curves in normalized target space.
(d) Total intensity $E(t)=\int |\hat \psi|^2 dx$.
(e) Complex-plane trajectory of $\hat \psi(x_0,t)$ at $x_0=0$.
(f) Peak amplitude $\max_x |\hat \psi(x,t)|$.
(g) Center of intensity $x_c(t)$.
All quantities in (a), (d)-(g) are computed from de-normalized predictions on the full space-time grid. All models are trained using the same optimization protocol: Adam with a learning rate of \(10^{-3}\), batch size 64, 64 randomly sampled space--time queries per trajectory per epoch, and a total of 1500 epochs, gradient clipping with a maximum norm of 1.0.}
\label{fig:nlse}
\end{figure*}

\begin{table*}[t]
\centering
\setlength{\tabcolsep}{5pt}
\renewcommand{\arraystretch}{1.12}
\caption{
Comparison on the NLSE benchmark.
Panel (a) reports the number of learnable parameters and the test-set
full-field complex MSE.
Panel (b) reports dataset-level trajectory errors in mass, momentum,
peak amplitude, and centre of mass for the five representative models
evaluated in the physical-fidelity study. For each quantity \(q\in\{M,P,A_{\mathrm{max}},x_c\}\), the trajectory
error for one test sample is defined as $\varepsilon_q=\frac{N_t^{-1}\sum_{n=1}^{N_t}\left|\widehat q(t_n)-q(t_n)\right|}{N_t^{-1}\sum_{n=1}^{N_t}\left|q(t_n)\right|}.$
All results are reported as the mean and standard deviation across three
independent seed-level test-set means.
Lower values are better.
}
\label{tab:full_results}
\small

\begin{tabular}{@{}lcc@{}}
\toprule
\multicolumn{3}{l}{\textbf{(a) Full-field reconstruction}}\\
\textbf{Model}
&
\textbf{Params (M)}
&
\textbf{Full-field MSE}
\\
\midrule
Vanilla
& 1.447
& \(1.242\times10^{-2}\pm1.065\times10^{-4}\)
\\
Concat
& 1.450
& \(5.202\times10^{-3}\pm9.443\times10^{-5}\)
\\
FiLM
& 1.846
& \(4.985\times10^{-3}\pm6.338\times10^{-5}\)
\\
OFormer adaptation
& 1.496
& \(7.059\times10^{-3}\pm2.871\times10^{-3}\)
\\
RG
& 1.587
& \(3.110\times10^{-3}\pm2.914\times10^{-5}\)
\\
MH-RG (\(R=6\))
& 1.464
& \(1.779\times10^{-3}\pm2.111\times10^{-4}\)
\\
MH-RG (\(R=12\))
& 1.473
& \(1.602\times10^{-3}\pm3.406\times10^{-5}\)
\\
\textbf{MH-RG (\(R=18\))}
& \textbf{1.482}
& \(\mathbf{1.406\times10^{-3}\pm1.854\times10^{-4}}\)
\\
MH-RG (\(R=24\))
& 1.491
& \(1.573\times10^{-3}\pm2.644\times10^{-4}\)
\\
\bottomrule
\end{tabular}

\vspace{0.8em}

\begin{tabular}{@{}lcccc@{}}
\toprule
\multicolumn{5}{l}{\textbf{(b) Dataset-level physical-fidelity errors}}\\
\textbf{Model}
&
\(\boldsymbol{\varepsilon_M}\)
&
\(\boldsymbol{\varepsilon_P}\)
&
\(\boldsymbol{\varepsilon_{A_\mathrm{max}}}\)
&
\(\boldsymbol{\varepsilon_{x_c}}\)
\\
\midrule
Vanilla
& \(0.1265\pm0.0133\)
& \(1.293\pm0.261\)
& \(0.0605\pm0.0062\)
& \(0.3116\pm0.0144\)
\\
FiLM
& \(0.0696\pm0.0112\)
& \(1.019\pm0.026\)
& \(0.0337\pm0.0012\)
& \(0.2125\pm0.0047\)
\\
OFormer adaptation
& \(0.0728\pm0.0454\)
& \(0.910\pm0.113\)
& \(0.0394\pm0.0159\)
& \(0.2049\pm0.0354\)
\\
RG
& \(0.0538\pm0.0045\)
& \(0.861\pm0.082\)
& \(0.0289\pm0.0016\)
& \(0.1830\pm0.0081\)
\\
\textbf{MH-RG (\(R=18\))}
& \(\mathbf{0.0286\pm0.0049}\)
& \(\mathbf{0.672\pm0.054}\)
& \(\mathbf{0.0203\pm0.0009}\)
& \(\mathbf{0.1094\pm0.0066}\)
\\
\bottomrule
\end{tabular}

\vspace{0.5em}

\begin{minipage}{0.98\textwidth}
\footnotesize
\end{minipage}
\end{table*}

The spatiotemporal maps in Fig.~\ref{fig:nlse}(a) show where these differences arise. The dominant discrepancies for all models are concentrated near the collision core and along the post-collision tails. Vanilla produces the largest errors in both amplitude and complex phase. Concat and FiLM recover the gross wavepacket structure more accurately, but residual errors remain visible in the interaction region. RG reduces these errors further, and MH-RG gives the cleanest reconstruction overall. The MSE boxplot and convergence curves in Fig.~\ref{fig:nlse}(b)-(c) show that the advantage of MH-RG is visible during overall performance and optimization.
The same conclusion is supported by the physical diagnostics in Fig.~\ref{fig:nlse}(d)-(g). Although no conservation law is imposed during training, MH-RG remains closest to the reference trajectory of the total intensity.
It also gives the most accurate peak-amplitude curve through the strong interaction window, the closest complex-plane trajectory at $x_0=0$, and the smallest drift in the center of intensity.
These quantities are derived from the predicted field and are therefore not independent training targets. Their agreement with the reference solution indicates that the gain is physically meaningful rather than merely pointwise.

 The dataset-level analysis for conditioned DeepONet families together with an external attention-based baseline are listed in Table \ref{tab:full_results}. Panel(a) reports the OFormer reduces the full-field error of Vanilla DeepONet from \(1.242\times10^{-2}\) to \(7.059\times10^{-3}\), a \(43.2\%\) reduction. Nevertheless, MH-RG (\(R=18\)) reaches \(1.406\times10^{-3}\), corresponding to an \(80.1\%\) reduction relative to OFormer at nearly identical model size. Panel (b) extends the comparison from field reconstruction to trajectory-level physical quantities. Mass and momentum measure the recovery of two conserved quantities of the NLSE, while peak amplitude and centre of mass quantify localization and
collective transport. MH-RG achieves the lowest error for all four quantities. This shows that its lower full-field MSE is accompanied by a more accurate reconstruction of the principal physical trajectories, rather than being
limited to a pointwise improvement in the complex field.

\begin{table*}[t]
\centering
\caption{
Controlled mechanism studies on the NLSE benchmark. Panel (a) uses the single-head dense RG architecture. Here, \(P\), \(B\), and \(T\) denote the pre-branch modulator, branch residual gate, and trunk residual gate, respectively; a subscript of \(1\) indicates that the corresponding module is enabled, whereas \(0\) indicates that it is disabled. Panel (b) uses the low-rank gate parameterization of MH-RG to separate the effects of low-rank compression, multiple readouts, and
head-specific conditioning. The learned-context and fixed-random-projection controls use the same head-specific \(R=18\) model and differ only in how the six-dimensional context vector is constructed.
}
\label{tab:mechanism_controls}
\small
\setlength{\tabcolsep}{4pt}
\renewcommand{\arraystretch}{1.08}
\begin{tabularx}{\textwidth}{
@{}p{0.27\textwidth}
X
>{\centering\arraybackslash}p{0.18\textwidth}@{}
}
\toprule
\textbf{Control configuration}
&
\textbf{Conditioning structure or context}
&
\textbf{Full-field MSE}
\\
\midrule

\multicolumn{3}{l}{
\textbf{(a) Complete \(2^3\) ablation of conditioning placement}
}
\\

\(\mathrm{P}_{0}\mathrm{B}_{0}\mathrm{T}_{0}\)
&
All three conditioning pathways disabled
&
\(1.242\times10^{-2}\)
\\

\(\mathrm{P}_{0}\mathrm{B}_{1}\mathrm{T}_{0}\)
&
Branch-side residual conditioning only
&
\(8.494\times10^{-3}\)
\\

\(\mathrm{P}_{0}\mathrm{B}_{0}\mathrm{T}_{1}\)
&
Trunk-side residual conditioning only
&
\(8.451\times10^{-3}\)
\\

\(\mathrm{P}_{0}\mathrm{B}_{1}\mathrm{T}_{1}\)
&
Branch- and trunk-side residual conditioning
&
\(6.912\times10^{-3}\)
\\

\(\mathrm{P}_{1}\mathrm{B}_{0}\mathrm{T}_{0}\)
&
Pre-branch modulation only
&
\(3.561\times10^{-3}\)
\\

\(\mathrm{P}_{1}\mathrm{B}_{1}\mathrm{T}_{0}\)
&
Pre-branch modulation with branch-side residual conditioning
&
\(3.394\times10^{-3}\)
\\

\(\mathrm{P}_{1}\mathrm{B}_{0}\mathrm{T}_{1}\)
&
Pre-branch modulation with trunk-side residual conditioning
&
\(3.405\times10^{-3}\)
\\

\(\mathrm{P}_{1}\mathrm{B}_{1}\mathrm{T}_{1}\)
&
All three conditioning pathways enabled
&
\(3.110\times10^{-3}\)
\\

\midrule

\multicolumn{3}{l}{
\textbf{(b) Low-rank multi-head and context-source controls}
}
\\

Single-readout low-rank control (\(R=1\))
&
Physical descriptors with one low-rank conditioned readout
&
\(3.253\times10^{-3}\)
\\

Shared-gate multi-readout control (\(R=18\))
&
Eighteen readouts sharing identical branch and trunk gates
&
\(2.003\times10^{-3}\)
\\

Head-specific multi-readout control (\(R=18\))
&
Head-dependent coefficient maps over shared low-rank gate bases
&
\(\mathbf{1.406\times10^{-3}}\)
\\

Learned-context control (\(R=18\))
&
Trainable \(256\!\to\!128\!\to\!6\) context encoder optimized jointly with the operator
&
\(3.463\times10^{-3}\)
\\

Fixed-random-projection control (\(R=18\))
&
Six-dimensional context obtained from a fixed Gaussian projection of the normalized initial field
&
\(5.444\times10^{-3}\)
\\

\bottomrule
\end{tabularx}
\end{table*}

The controlled studies in Table~\ref{tab:mechanism_controls}
separate the effects of conditioning placement, residual modulation,
multi-head decomposition, and context construction.
Panel~(a) identifies the pre-branch modulator as the dominant individual
conditioning pathway.
Relative to the fully unconditioned control
\(\mathrm{P}_{0}\mathrm{B}_{0}\mathrm{T}_{0}\),
pre-branch modulation alone,
\(\mathrm{P}_{1}\mathrm{B}_{0}\mathrm{T}_{0}\),
reduces the MSE by \(71.3\%\).
Branch-side and trunk-side residual conditioning alone reduce the MSE by
\(31.6\%\) and \(32.0\%\), respectively, while their joint use without
pre-branch modulation reduces it by \(44.3\%\).
When pre-branch modulation is already active, adding both latent residual
pathways further reduces the MSE from
\(3.561\times10^{-3}\) to \(3.110\times10^{-3}\),
corresponding to an additional \(12.7\%\) improvement.
These results show that the pre-branch pathway provides the largest
individual gain, while the branch and trunk residual gates provide
additional latent-space refinement.
Panel~(b) examines the multi-head mechanism and the source of the compact
context vector.
Moving from the single-readout low-rank control to the shared-gate
multi-readout control reduces the MSE by \(38.4\%\), even though all
readouts use the same conditioned latent interaction.
This improvement is therefore attributed to the optimization and
reparameterization associated with multiple additive readouts.
Allowing the descriptor-to-gate maps to become head dependent reduces the
shared-gate error by a further \(29.8\%\), and by \(56.8\%\) relative to the
single-readout model.
This additional reduction provides direct evidence that non-identical
head-dependent gates increase the conditional expressivity of the model.
The learned-context control replaces the handcrafted physical descriptors
with a trainable encoder that maps the initial-field sensor vector to a six-dimensional context.
This encoder is optimized jointly with the head-specific operator.
The fixed-random-projection control provides a non-physical,
non-trainable six-dimensional compression of the same normalized
initial-field sensor vector.
A Gaussian random projection from the original \(256\)-dimensional
sensor space to six dimensions is sampled once before training and then
kept fixed throughout the experiment.
The projected components are standardized using statistics computed
from the training set before being supplied to the conditioning
pathways.
Because neither the projection nor the resulting context is learned,
this control preserves the context dimensionality while removing both
physical interpretation and trainable feature extraction.
Using the same head-specific model, the physical,
learned, and fixed-random-projection contexts yield MSEs of
\(1.406\times10^{-3}\),
\(3.463\times10^{-3}\), and
\(5.444\times10^{-3}\), respectively.
The learned-context result shows that the conditioning architecture can
extract useful compact information directly from the initial field.
However, the lower error obtained with the physical descriptors indicates
that physically selected observables provide a stronger inductive bias in
the available-data regime.
The fixed-random-projection result further shows that the improvement
cannot be explained merely by supplying an arbitrary six-dimensional
compression of the initial condition.

\begin{figure*}[h!]
\centering
\includegraphics[width=0.95\textwidth]{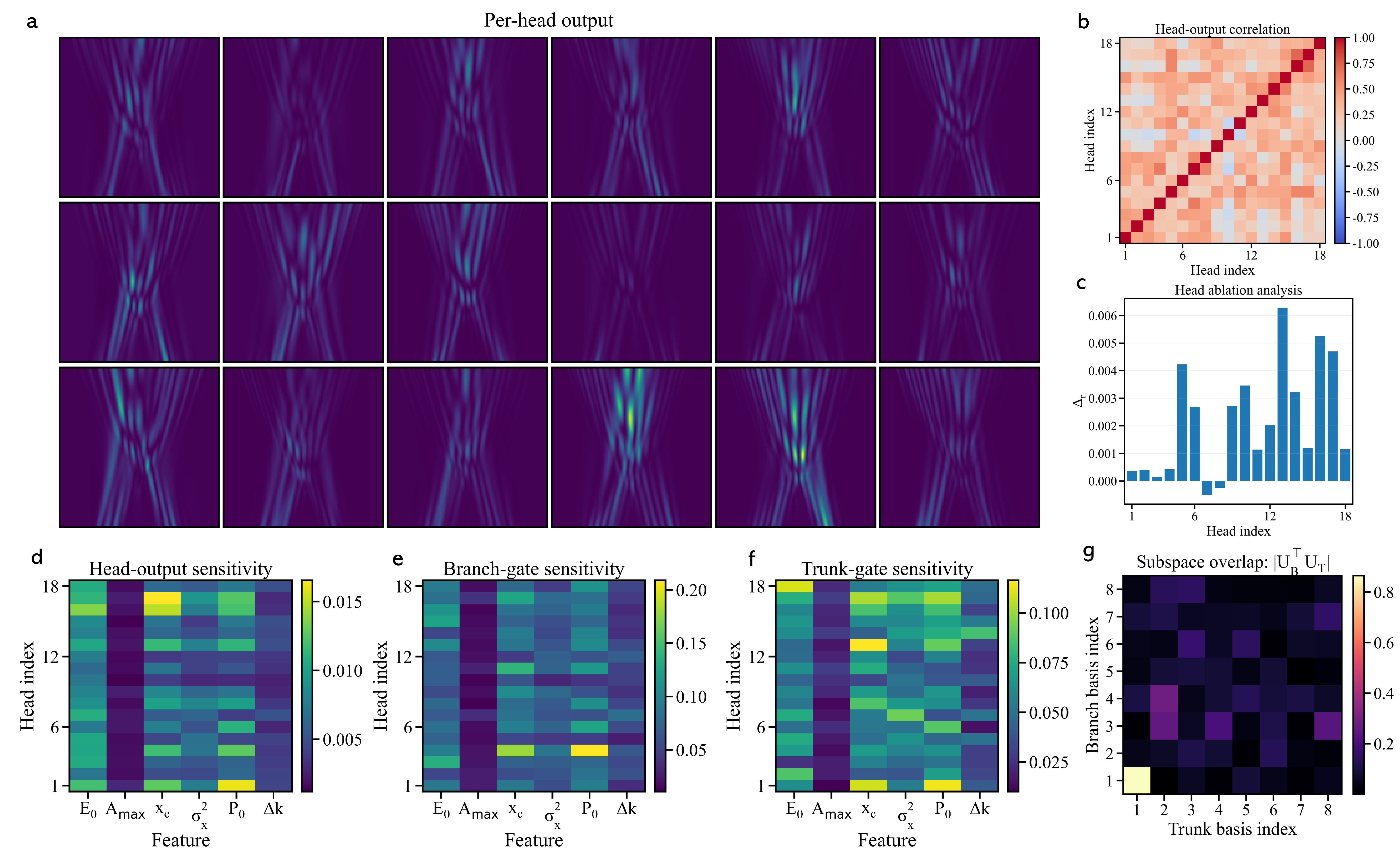}
\caption{\textbf{Mechanistic analysis of the MH-RG model on the NLSE benchmark.}
(a) Per-head output components for a representative test trajectory.
(b) Head-output correlation matrix on the representative sample. 
(c) Head-ablation analysis, where the leave-one-head-out effect is defined as
\(\Delta_r=\mathrm{MSE}(\widehat{\psi}^{(-r)},\psi)-\mathrm{MSE}(\widehat{\psi},\psi)\),
with \(\widehat{\psi}^{(-r)}\) denoting the prediction obtained after removing head \(r\). Positive values indicate that the corresponding head improves the full-field prediction.
(d) Sensitivity of each head output to the six initial-state descriptors.
(e) Sensitivity of the branch-side gates to the same descriptors.
(f) Sensitivity of the trunk-side gates to the same descriptors.
(g) Overlap map between the learned low-rank branch and trunk gate subspaces, computed from the shared upsampling matrices.}
\label{fig:machanism}
\end{figure*}

\begin{figure*}[h!]
\centering
\includegraphics[width=0.95\textwidth]{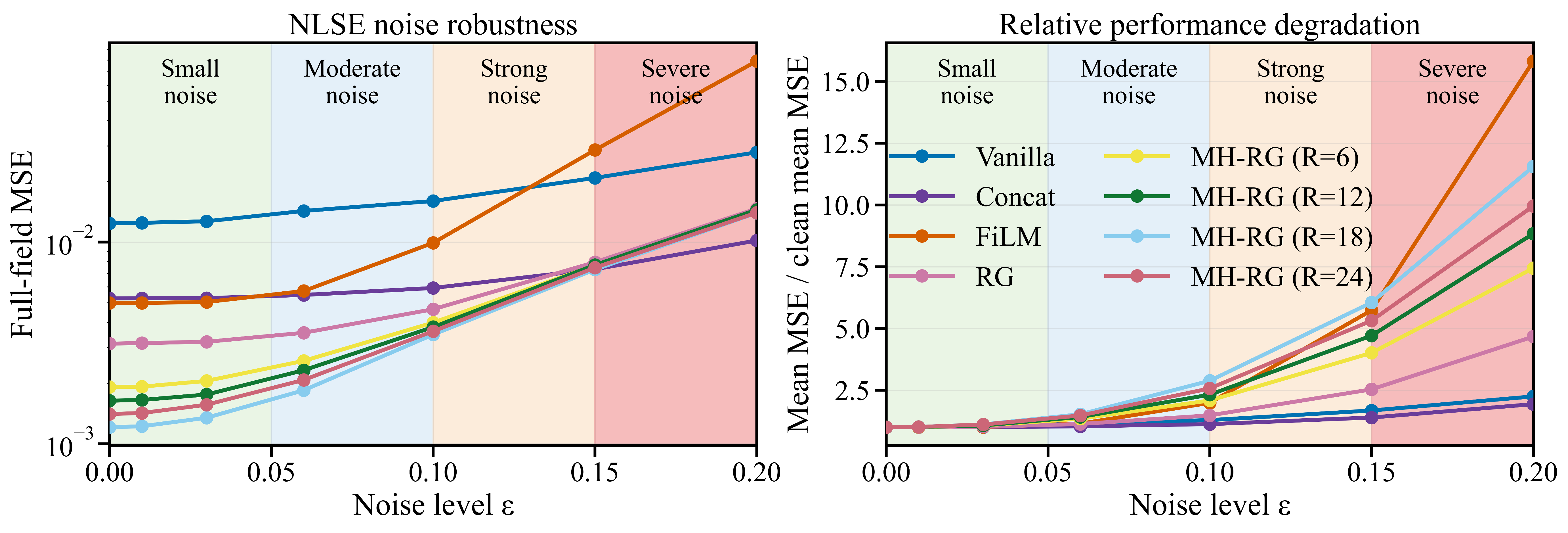}
\caption{\textbf{Noise robustness on the NLSE benchmark.}
Complex Gaussian noise is added to the observed initial condition at inference time, with perturbation level \(\epsilon\). The same noisy input is used for both the sensor representation and the extracted physical descriptors. Shaded regions indicate small, moderate, strong, and severe noise regimes.
\textbf{Left:} Test-set full-field MSE versus noise level.
\textbf{Right:} Relative degradation, measured by the ratio between the noisy-input mean MSE and the clean-input mean MSE of each model.}
\label{fig:nlse_noise}
\end{figure*}

To examine how the multi-head mechanism operates, we further analyze the MH-RG
(\(R=18\)) model in Fig.~\ref{fig:machanism}.
The per-head outputs in Fig.~\ref{fig:machanism}(a) share the same broad
collision and propagation structure, but assign different relative weights
to particular space--time regions.
This indicates that the additive heads do not simply reproduce identical
copies of the full prediction.
The head-output correlation matrix in
Fig.~\ref{fig:machanism}(b) provides a direct comparison of these
representative outputs.
The off-diagonal correlations are generally intermediate rather than being
uniformly close to either zero or one.
Thus, the heads remain coordinated through the common branch--trunk
representation, while retaining non-identical output patterns.
The representative leave-one-head-out analysis in
Fig.~\ref{fig:machanism}(c) further shows that the effect of removing a head
is strongly non-uniform.
Most heads increase the prediction error when removed, whereas a small
number have near-zero or slightly negative effects for this particular
trajectory.
Such sample-level variation is expected in a jointly optimized additive
decomposition, since the heads are not trained independently and their
contributions can overlap or partially compensate one another for a given
input. To examine whether this behavior persists across the test set, we
evaluated the same quantities over all \(200\) test trajectories and all
three trained seeds. The mean off-diagonal correlation between head outputs
ranges from \(0.234\) to \(0.264\), indicating that the heads are neither
independent nor collapsed into redundant copies. Although removing an
individual head may have a negligible or slightly negative effect for a
particular trajectory, removing any head increases the dataset-averaged
MSE. This shows that the heads make complementary contributions at the
dataset level, even though their effects can vary from sample to sample.

The sensitivity maps in Fig.~\ref{fig:machanism}(d)--(f) provide a
complementary view of the conditioned decomposition.
Different heads exhibit different sensitivities to the six physical
descriptors at the level of the final head output and at the branch and trunk gates.
These maps show that the descriptor-conditioned modulation is
distributed non-uniformly across the heads. They support non-identical,
head-dependent use of the physical context.
Finally, the subspace-overlap matrix in
Fig.~\ref{fig:machanism}(g) compares the learned branch-side and trunk-side
low-rank bases.
The overlap is concentrated in several basis-direction pairs rather than
being either uniformly large or uniformly negligible.
The two gate subspaces therefore share selected directions while retaining
distinct components.
This is consistent with the intended structure of the model: branch and
trunk heads are coupled through compact low-rank representations, but are
not constrained to use identical modulation directions.

We further test robustness by perturbing the observed NLSE initial condition at inference time, without retraining and without modifying the reference solution. The noisy input is defined as
\begin{equation}
u_0^{\mathrm{noisy}}(x)
=
u_0(x)
+
\epsilon \,\mathrm{rms}(u_0)\,\xi(x),
\end{equation}
where \(\epsilon\) controls the perturbation strength and \(\xi\) is unit-variance complex Gaussian noise and $\mathrm{rms}(u_0)$ denotes the root-mean-square amplitude of the initial field. Both the sensor representation and the physical descriptors are computed from \(u_0^{\mathrm{noisy}}\), so the test probes corruption of both the raw field observation and the conditioning signal. As shown in Fig.~\ref{fig:nlse_noise}, the residual-gated models preserve the lowest or near-lowest absolute full-field MSE throughout the small- and moderate-noise regimes. Under stronger corruption, the multi-head models degrade more rapidly relative to their clean counterparts, whereas Vanilla and Concat appear more stable in relative terms but remain less accurate overall. This indicates that structured gating improves practical accuracy under moderate observational uncertainty, while heavy corruption increasingly limits the benefit of descriptor-based modulation.

As a supportive study for the architecture, we varied the pre-branch gate amplitude $\alpha_{\mathrm{pre}}$ while fixing the latent branch- and trunk-side gate amplitudes to $\alpha_B=\alpha_T=0.5$. As shown in Table~\ref{tab:alpha_sensitivity}, the error decreases substantially when increasing $\alpha_{\mathrm{pre}}$ from $0.5$ to $1.0$, and then increases slightly at $\alpha_{\mathrm{pre}}=1.5$. These results indicate a clear preference for a moderate pre-branch amplitude near \(\alpha_{\mathrm{pre}}=1\) within the tested range.This trend is consistent with the structure of the residual pre-branch gate. When $\alpha_{\mathrm{pre}}<1$, the gate can attenuate sensor channels but cannot fully suppress them. At $\alpha_{\mathrm{pre}}=1$, channel-wise modulation can reach zero, allowing sharper input-side selection. For $\alpha_{\mathrm{pre}}>1$, the gate may become negative on some channels, introducing sign changes that appear not to be beneficial in this setting.
Second, we fixed $\alpha_{\mathrm{pre}}=1.0$ and varied the latent gate amplitudes jointly as $\alpha_B=\alpha_T$. The corresponding results are also reported in Table~\ref{tab:alpha_sensitivity}. The best performance is obtained at $\alpha_B=\alpha_T=0.5$, while larger values do not provide further gains.
The latent gates therefore exhibit a different sensitivity profile from the pre-branch gate. Their role is better interpreted as a residual reweighting of learned embeddings rather than an input-level masking mechanism. For
$\gamma(\phi)=1+\alpha\tanh(f(\phi))$,
the channel-wise scaling lies in $[1-\alpha,\,1+\alpha]$. Thus, when $\alpha=0.5$, the modulation remains in $[0.5,1.5]$ and stays non-negative, corresponding to a moderate reshaping of the latent representation. 

\begin{table}[h]
\centering
\caption{Sensitivity of MH-RG ($R=18$) to the gate factors on the NLSE benchmark. Reported values are test-set full-field MSE aggregated over three independent random seeds (\(\mathrm{mean}\pm\mathrm{std}\) across seed means). Lower is better.}
\label{tab:alpha_sensitivity}
\setlength{\tabcolsep}{10pt}
\renewcommand{\arraystretch}{1.12}
\begin{tabular}{cc}
\toprule
\multicolumn{2}{c}{\textbf{(a) Pre-branch gate sweep with $\alpha_B=\alpha_T=0.5$}} \\
\midrule
\textbf{$\alpha_{\mathrm{pre}}$} & \textbf{Full-field MSE} \\
\midrule
0.5 & $2.133 \times 10^{-3} \pm 1.956 \times 10^{-4}$ \\
\textbf{1.0} & $\mathbf{1.406 \times 10^{-3} \pm 1.854 \times 10^{-4}}$ \\
1.5 & $1.505 \times 10^{-3} \pm 7.057 \times 10^{-5}$ \\
\midrule
\multicolumn{2}{c}{\textbf{(b) Latent gate sweep with $\alpha_{\mathrm{pre}}=1.0$ and $\alpha_B=\alpha_T$}} \\
\midrule
\textbf{$\alpha_B=\alpha_T$} & \textbf{Full-field MSE} \\
\midrule
0.1 & $1.752 \times 10^{-3} \pm 9.358 \times 10^{-5}$ \\
\textbf{0.5} & $\mathbf{1.406 \times 10^{-3} \pm 1.854 \times 10^{-4}}$ \\
1.0 & $1.438 \times 10^{-3} \pm 9.947 \times 10^{-4}$ \\
\bottomrule
\end{tabular}
\end{table}

We further performed a temporal-extrapolation test in which MH-RG (R=18) and OFormer
were trained only on \(t\leq1.5\) and evaluated on the unseen interval
\(1.5<t\leq2.0\). Averaged over three independent seeds, MH-RG gives an
extrapolation MSE of \(1.062\times10^{-1}\pm1.160\times10^{-2}\),
compared with \(2.960\times10^{-1}\pm5.919\times10^{-2}\) for OFormer. These results indicate
that the strong interpolation advantage of MH-RG does not directly
translate into uniformly improved temporal extrapolation. Long-time
prediction beyond the training horizon therefore remains an open
limitation.

Overall, the NLSE benchmark shows that compact initial-state descriptors are useful for operator learning in phase-sensitive coherent wave dynamics, and their largest individual benefit is obtained through descriptor-conditioned pre-branch modulation, with bounded latent gates supplying additional refinement. Meanwhile, the multi-head extension improves further over the RG model while remaining lightweight in parameter count. The comparison with OFormer establishes that this gain is not restricted to a weak DeepONet baseline, while the learned-context, shared-gate multi-readout, and factorial controls identify which parts of the architecture are responsible for the improvement.

\subsection{Generalization to Dissipative Trapped Systems: 2D Shift-Breathers}

We next introduce a higher-dimensional dissipative system in which the initial condition carries both spatial displacement and phase tilt. The evolution is governed by the two-dimensional damped Gross-Pitaevskii equation~\cite{Choi1998PRA, Reeves2012PRA}:
\begin{equation}
\label{eq:gpe2d_sb}
\begin{aligned}
i\,\partial_t \psi(x,y,t)
&=
-\frac{1}{2}\,\Delta\psi(x,y,t)
+V(x,y)\,\psi(x,y,t) \\
&\quad
+g\,|\psi(x,y,t)|^2\psi(x,y,t)
-i\,\gamma\,\psi(x,y,t),
\end{aligned}
\end{equation}
where $\Delta=\partial_{xx}+\partial_{yy}$, $g=1.0$, and $\gamma=0.05$. The external potential is the isotropic harmonic trap
\begin{equation}
V(x,y)=\frac{1}{2}\,\Omega^2(x^2+y^2),
\qquad \Omega=1.5.
\end{equation}
The computational domain is $(x,y)\in[-4,4]^2$, discretized on a uniform $128\times128$ grid, and the trajectories are computed on $t\in[0,2]$ with $\Delta t=0.01$ using a split-step Fourier solver.

Compared with the radial damped-breather setting, we consider a shifted and phase-tilted Gaussian initial state,
\begin{equation}
\label{eq:gpe2d_sb_ic}
\begin{aligned}
\psi_0(x,y)
&=
A\exp\!\left[
-\frac{(x-x_0)^2+(y-y_0)^2}{2\sigma^2}
\right] \\
&\quad\times
\exp\!\left[i(k_xx+k_yy+\phi_0)\right].
\end{aligned}
\end{equation}
with
\[
A\in[1.0,1.5],\qquad
\sigma\in[0.8,1.3],\qquad
\phi_0\in[0,2\pi),
\]
\[
x_0,y_0\in[-0.8,0.8],\qquad
k_x,k_y\in[-0.8,0.8].
\]
This family is physically more informative because the trap supports both center-of-mass oscillation and breathing oscillation, and the displaced, tilted initial state couples the two. 
With damping, the total norm gradually decreases while the displaced packet undergoes trapped translational and breathing dynamics.
The resulting dynamics are therefore more structured than a simple radial breather and provide a more challenging test for conditioned neural operators. We generate $200$ training trajectories and $50$ test trajectories.

From each initial condition we extract a ten-dimensional descriptor vector
\[
\phi=
[E_0,\;A_{\max},\;x_c,\;y_c,\;\sigma_x^2,\;\sigma_y^2,\;P_x,\;P_y,\;\Delta k_x,\;\Delta k_y].
\]
These quantities are defined by
\begin{equation}
E_0=\iint |\psi_0(x,y)|^2\,dx\,dy,
\qquad
A_{\max}=\max_{x,y}|\psi_0(x,y)|,
\end{equation}
\begin{equation}
x_c=\frac{\iint x\,|\psi_0|^2\,dx\,dy}{E_0},
\qquad
y_c=\frac{\iint y\,|\psi_0|^2\,dx\,dy}{E_0},
\end{equation}
\begin{equation}
\sigma_x^2=\frac{\iint (x-x_c)^2|\psi_0|^2\,dx\,dy}{E_0},
\qquad
\sigma_y^2=\frac{\iint (y-y_c)^2|\psi_0|^2\,dx\,dy}{E_0}.
\end{equation}
The remaining four descriptors are computed from the Fourier power spectrum $|\tilde{\psi}_0(k_x,k_y)|^2$,
\begin{equation}
P_x=
\frac{\iint k_x\,|\tilde{\psi}_0|^2\,dk_x\,dk_y}
{\iint |\tilde{\psi}_0|^2\,dk_x\,dk_y},
\qquad
P_y=
\frac{\iint k_y\,|\tilde{\psi}_0|^2\,dk_x\,dk_y}
{\iint |\tilde{\psi}_0|^2\,dk_x\,dk_y},
\end{equation}

\begin{equation}
\Delta k_x=
\left[
\frac{\iint (k_x-P_x)^2|\tilde{\psi}_0|^2\,dk_x\,dk_y}
{\iint |\tilde{\psi}_0|^2\,dk_x\,dk_y}
\right]^{1/2},
\end{equation}
\begin{equation}
\Delta k_y=
\left[
\frac{\iint (k_y-P_y)^2|\tilde{\psi}_0|^2\,dk_x\,dk_y}
{\iint |\tilde{\psi}_0|^2\,dk_x\,dk_y}
\right]^{1/2}.
\end{equation}
These ten descriptors summarize total intensity, localization, displacement, momentum, and spectral spread of the initial condensate.

The training protocol follows the same principle used in the 1D NLSE benchmark. Relative to the 1D setting, the 2D branch input grows from $2N_x$ to $2N_xN_y$, while the trunk input grows from $(x,t)$ to $(x,y,t)$. To limit rapid parameter growth, the pre-branch residual modulator is implemented in low-rank form. All models are trained with Adam and a StepLR scheduler. The latter is introduced to improve optimization stability in the late stage of training, where the loss often enters a broad plateau accompanied by noticeable oscillations. Reducing the learning rate in stages helps damp these fluctuations and promotes more stable convergence.

\begin{table}[t]
\centering
\setlength{\tabcolsep}{5pt}
\caption{Model comparison on the 2D damped Gross-Pitaevskii shift-breather benchmark. All models use the same global-normalization protocol and are trained with Adam and StepLR. We report the number of learnable parameters and the test-set full-field MSE aggregated over three independent random seeds (\(\mathrm{mean}\pm\mathrm{std}\) across seed means). Lower MSE is better.}
\label{tab:gpe2d_results}
\begin{tabular}{@{}l c c@{}}
\toprule
\textbf{Model} & \textbf{Params (M)} & \textbf{Full-field MSE} \\
\midrule
Vanilla                   & 8.719 & $1.188 \times 10^{-2} \pm 2.978 \times 10^{-4}$ \\
Concat                    & 8.722 & $1.165 \times 10^{-2} \pm 5.714 \times 10^{-4}$ \\
FiLM                      & 8.921 & $5.976 \times 10^{-3} \pm 1.742 \times 10^{-4}$ \\
RG                        & 8.854 & $4.540 \times 10^{-3} \pm 2.495 \times 10^{-4}$ \\
MH-RG (\(R=2\))           & 8.829 & $3.650 \times 10^{-3} \pm 8.398 \times 10^{-5}$ \\
MH-RG (\(R=4\))           & 8.837 & $3.180 \times 10^{-3} \pm 1.566 \times 10^{-4}$ \\
MH-RG (\(R=8\))           & 8.852 & $3.038 \times 10^{-3} \pm 1.825 \times 10^{-4}$ \\
\textbf{MH-RG (\(R=16\))} & \textbf{8.882} & $\mathbf{3.004 \times 10^{-3} \pm 2.163 \times 10^{-4}}$ \\
\bottomrule
\end{tabular}
\end{table}

The error report in Table~\ref{tab:gpe2d_results} shows that simply appending the physical descriptors to the input is not enough. FiLM improves substantially, which indicates that conditional modulation is useful. RG improves further. The best results are obtained by the multi-head residual-gated family, and within the tested range the performance improves with increasing head count, but the improvement gradually saturates for larger $R$. 

\begin{figure*}[t]
\centering
\includegraphics[width=0.98\textwidth]{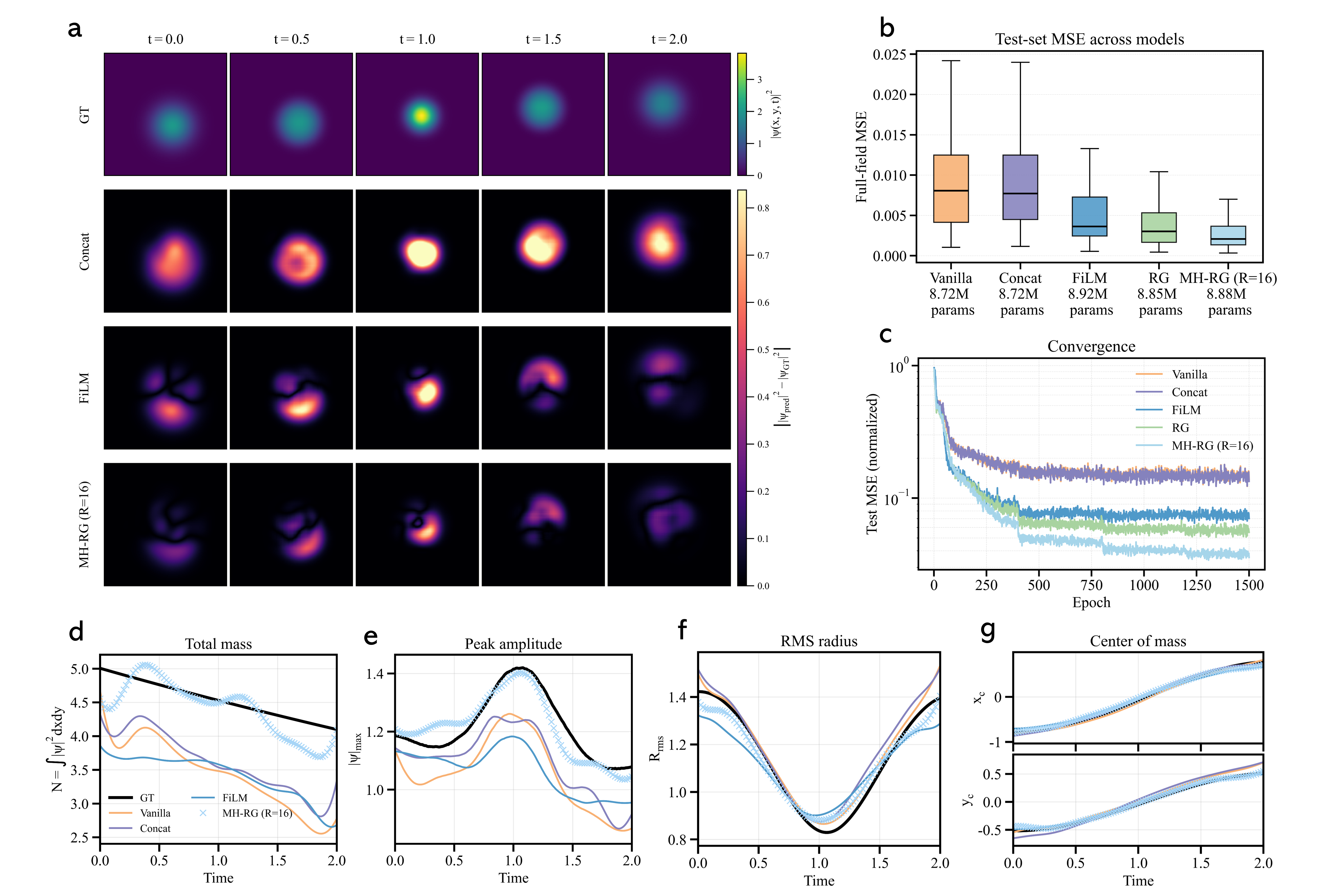}
\caption{\textbf{2D damped Gross-Pitaevskii shift-breather benchmark.}
\textbf{(a)} Representative density snapshots and density-error maps for one test trajectory. The first row shows the ground-truth density \(| \psi_{GT}(x,y,t)|^2\). The remaining rows show \(\bigl||\hat \psi_{\mathrm{pred}}|^2-|\psi_{\mathrm{GT}}|^2\bigr|\) for Concat, FiLM, and MH-RG (\(R=16\)) at selected times.
\textbf{(b)} Test-set full-field MSE across models.
\textbf{(c)} Mean convergence of the normalized test loss across seeds.
\textbf{(d-g)} Physical diagnostics computed from the reconstructed full field: total mass
$N(t)=\iint |\hat \psi(x,y,t)|^2\,dx\,dy,$
peak amplitude, RMS radius $
R_{\mathrm{rms}}(t)=
\left[
\frac{\iint (x^2+y^2)|\hat \psi(x,y,t)|^2\,dx\,dy}
{\iint |\hat \psi(x,y,t)|^2\,dx\,dy}
\right]^{1/2},
$
and the center-of-mass coordinates.
}
\label{fig:gpe2d_shiftbreather}
\end{figure*}

Figure~\ref{fig:gpe2d_shiftbreather}(a) shows that the representative errors are not confined to isolated pixels. Concat fails to reconstruct the displaced packet shape and produces broad structural intensity errors across the entire cycle. FiLM recovers the coarse motion more accurately, but visible shape distortion remains. MH-RG suppresses the error further, especially near the compression stage where the packet is both spatially localized and dynamically shifted. 
The convergence curves in Fig.~\ref{fig:gpe2d_shiftbreather}(c) show a similarly informative trend. After each StepLR decay, the test loss of FiLM, RG, and MH-RG decreases further, whereas Vanilla and Concat flatten at substantially higher error levels. This contrast suggests that physically meaningful descriptors, when integrated through explicit modulation, provide a more effective conditioned representation that remains refinable in the low-learning-rate regime. 

The physical diagnostics in Fig.~\ref{fig:gpe2d_shiftbreather}(d-g) show that the accuracy improvement is physically meaningful. Since the system is damped, the relevant test is whether the model reproduces the correct decay trajectory. The same holds for the peak amplitude and the RMS radius, which together characterize the breathing cycle. The center-of-mass coordinates track the trapped translational motion. MH-RG follows the reference trajectories more closely, in particular, it better captures the coupled evolution of displacement and compression, which is the defining feature of this benchmark.

This study sharpens the interpretation suggested by the NLSE results. In the conservative collision problem, structured conditioning improves prediction in a phase-sensitive nonlinear wave regime. In the present trapped dissipative system, the design remains effective when the dynamics are governed by coupled translation, breathing, and decay.

\section{Discussion}
\label{sec:discussion}

In this work, we showed that DeepONet can make more effective use of a small set of physically meaningful descriptors when they are introduced through structured modulation. The ablation study clarifies the role of each component. The pre-branch pathway provides the largest individual improvement by modifying how the initial state is encoded, while the branch and trunk residual gates provide additional latent and coordinate-dependent refinement. Their bounded, identity-centred form allows the descriptor pathway to act as a controlled correction to the original representation.
The multi-head extension improves the model in a different way. It increases conditional expressivity by combining several low-rank, descriptor-conditioned response channels without a large increase in parameter count. The head-count sweep, the shared-gate control, and the dataset-level leave-one-head-out analysis consistently show that the gain is not explained by additional readouts alone. The head-dependent gates provide further improvement, while the individual heads remain coordinated and contribute complementary corrections to the reconstructed solution.
The external comparison leads to a complementary conclusion. The OFormer adaptation is substantially better than Vanilla DeepONet, confirming that it is a meaningful attention-based baseline, yet MH-RG achieves an approximately fivefold lower full-field MSE at nearly identical model size.

A limitation of the present framework is that its performance depends on the choice of descriptors. The model is most suitable for systems whose dynamics can be summarized by a small number of physically meaningful quantities. These descriptors are not universal and must be selected separately for each PDE. Appropriate choices may include conserved quantities, balance-law variables, collective coordinates, localization measures, and spectral statistics. The learned-context experiment shows that the architecture can still operate without handcrafted descriptors, although the lower accuracy also demonstrates the benefit of using physically informed variables. For fully developed turbulence, broadband chaotic dynamics, or strongly forced multiscale systems, a compact set of static descriptors may be difficult to identify or may not contain enough information. In such cases, the conditioning information may need to be learned from the field, inferred from several observations in time, or updated dynamically during the evolution.

A further limitation concerns extension to three-dimensional fields. In the present implementation, the full field is flattened and encoded by a dense branch network, so the parameter and memory cost of the sensor-to-latent mapping grows directly with the number of spatial grid points. This becomes prohibitive for high-resolution 3D fields. A practical extension would therefore replace the dense branch with a structured spatial encoder, such as a convolutional, spectral, or scalable attention-based encoder, while retaining the descriptor-conditioned residual and multi-head mechanisms in the resulting latent space.

The temporal extrapolation limitation also suggests several natural directions for future work. One is to move beyond static conditioning based only on the initial state and instead incorporate dynamically updated or recurrent observables that evolve together with the solution. Another is to investigate whether the same design principle can be transferred beyond DeepONet to other neural operator backbones, including grid-based spectral and attention-based architectures, and whether it can improve predictive accuracy without sacrificing physical interpretability or substantially increasing the computational cost.

% ====== Insert before \printcredits ======
\section*{Declaration of Competing Interest}
The authors declare that they have no known competing financial interests or personal relationships that could have appeared to influence the work reported in this paper.

\section*{Data and Code Availability}
The data supporting the findings of this study will be made openly available upon acceptance of the manuscript. Additional implementation details, mathematical formulations, and architectural specifications are documented in the manuscript.
% =========================================


\begin{thebibliography}{99}

% ===== References cited in the Introduction =====
\bibitem{Agrawal2019}
G.~P. Agrawal.
\emph{Nonlinear Fiber Optics} (6th ed.).
Academic Press (2019).

\bibitem{Hasegawa1995}
A.~Hasegawa and Y.~Kodama.
\emph{Solitons in Optical Communications}.
Oxford University Press (1995).

\bibitem{DelHaye2007}
P.~Del'Haye, A.~Schliesser, O.~Arcizet, T.~Wilken, R.~Holzwarth, and T.~J. Kippenberg.
Optical frequency comb generation from a monolithic microresonator.
\emph{Nature} \textbf{450}, 1214--1217 (2007).

\bibitem{Kippenberg2018}
T.~J. Kippenberg, A.~L. Gaeta, M.~Lipson, and M.~L. Gorodetsky.
Dissipative Kerr solitons in optical microresonators.
\emph{Science} \textbf{361}(6402), eaan8083 (2018).

\bibitem{Fan2022}
Z.~Fan, D.~N. Puzyrev, and D.~V. Skryabin.
Topological soliton metacrystals.
\emph{Communications Physics} \textbf{5}, 248 (2022).

\bibitem{NA2023}
N~Amiune, Z.~Fan, V.~V. Pankratov, D.~N. Puzyrev, D.~V. Skryabin, K.~T. Zawilski, P.~G. Schunemann, and I.~Breunig.
Mid-infrared frequency combs and staggered spectral patterns in $\chi(2)$ microresonators.
\emph{Optics Express} \textbf{31}, 907-915 (2023).

\bibitem{Strecker2002}
K.~E. Strecker, G.~B. Partridge, A.~G. Truscott, and R.~G. Hulet.
Formation and propagation of matter-wave soliton trains.
\emph{Nature} \textbf{417}, 150--153 (2002).

\bibitem{Khaykovich2002}
L.~Khaykovich, F.~Schreck, G.~Ferrari, T.~Bourdel, J.~Cubizolles, L.~D. Carr, Y.~Castin, and C.~Salomon.
Formation of a matter-wave bright soliton.
\emph{Science} \textbf{296}, 1290--1293 (2002).

\bibitem{Nguyen2017}
J.~H. Nguyen, D.~Luo, and R.~G. Hulet.
Formation of matter-wave soliton trains by modulational instability.
\emph{Science} \textbf{356}, 422-426 (2017).

\bibitem{Karniadakis2021Review}
G.~E. Karniadakis, I.~G. Kevrekidis, L.~Lu, P.~Perdikaris, S.~Wang, and L.~Yang.
Physics-informed machine learning.
\emph{Nature Reviews Physics} \textbf{3}(6), 422--440 (2021).

\bibitem{Kovachki2023JMLR}
N.~Kovachki, Z.~Li, B.~Liu, K.~Azizzadenesheli, K.~Bhattacharya, A.~M. Stuart, and A.~Anandkumar.
Neural Operator: Learning Maps Between Function Spaces With Applications to PDEs.
\emph{Journal of Machine Learning Research} \textbf{24}(89), 1--97 (2023).

\bibitem{Raissi2019}
M.~Raissi, P.~Perdikaris, and G.~E. Karniadakis.
Physics-informed neural networks: A deep learning framework for solving forward and inverse problems involving nonlinear partial differential equations.
\emph{Journal of Computational Physics} \textbf{378}, 686--707 (2019).

\bibitem{Wang2021GradientPathologies}
S.~Wang, Y.~Teng, and P.~Perdikaris.
Understanding and mitigating gradient flow pathologies in physics-informed neural networks.
\emph{SIAM Journal on Scientific Computing} \textbf{43}(5), A3055--A3081 (2021).

\bibitem{Wang2022AdaptivePINN}
S.~Wang, X.~Yu, and P.~Perdikaris.
When and why PINNs fail to train: A neural tangent kernel perspective.
\emph{Journal of Computational Physics} \textbf{449}, 110768 (2022).

\bibitem{Ji2021Stiff}
W.~Ji, W.~Qiu, Z.~Shi, S.~Pan, and S.~Deng.
Stiff-PINN: Physics-Informed Neural Network for Stiff Chemical Kinetics.
\emph{Journal of Physical Chemistry A} \textbf{125}(36), 8098--8106 (2021).

\bibitem{lu2021deeponet}
L.~Lu, P.~Jin, G.~Pang, Z.~Zhang, and G.~E. Karniadakis.
Learning nonlinear operators via DeepONet based on the universal approximation theorem of operators.
\emph{Nature Machine Intelligence} \textbf{3}, 218--229 (2021).

\bibitem{Li2021FNO}
Z.~Li, N.~Kovachki, K.~Azizzadenesheli, B.~Liu, K.~Bhattacharya, A.~M. Stuart, and A.~Anandkumar.
Fourier Neural Operator for Parametric Partial Differential Equations.
\emph{International Conference on Learning Representations (ICLR)} (2021).

\bibitem{Li2022PINO}
Z.~Li, H.~Zheng, N.~Kovachki, D.~Jin, H.~Chen, B.~Liu, K.~Azizzadenesheli, and A.~Anandkumar.
Physics-informed neural operator for learning partial differential equations.
\emph{International Conference on Learning Representations (ICLR)} (2022).

\bibitem{Hao2023ICML}
Z.~Hao, Z.~Wang, H.~Su, C.~Ying, Y.~Dong, S.~Liu, Z.~Cheng, J.~Song, and J.~Zhu.
GNOT: A general neural operator transformer for operator learning.
\emph{International Conference on Machine Learning (ICML)} (2023).


\bibitem{Li2023OFormer}
Z.~Li, K.~Meidani, and A.~B.~Farimani.
Transformer for Partial Differential Equations' Operator Learning.
\emph{Transactions on Machine Learning Research (TMLR)} (2023).

\bibitem{Liu2023INO}
N.~Liu, Y.~Yu, H.~You, and N.~Tatikola.
INO: Invariant Neural Operators for Learning Complex Physical Systems with Momentum Conservation.
In \emph{Proceedings of The 26th International Conference on Artificial Intelligence and Statistics (AISTATS)}
(PMLR \textbf{206}), 6822--6838 (2023).


\bibitem{Zhang2024PIANO}
R.~Zhang, Q.~Meng and Z.~Ma.
Deciphering and integrating invariants for neural operator learning with various physical mechanisms.
\emph{National Science Review} \textbf{11(4): nwad336} (2024).

\bibitem{Lei2024LongTime}
G.~Lei, Z.~Lei, and L.~Shi.
Long-time Integration of Nonlinear Wave Equations with Neural Operators.
\emph{arXiv preprint arXiv:2410.15617} (2024).

\bibitem{lu2021deeponet_fair}
L.~Lu, X.~Meng, S.~Cai, Z.~Mao, S.~Goswami, Z.~Zhang, and G.~E. Karniadakis.
A comprehensive and fair comparison of two neural operators (with practical extensions) based on FAIR data.
\emph{Computer Methods in Applied Mechanics and Engineering} \textbf{393}, 114778 (2022).


\bibitem{Samuel2023}
S.~Lanthaler, R.~Molinaro, P.~Hadorn, and S. Mishra.
Nonlinear reconstruction for operator learning of PDEs with discontinuities.
\emph{International Conference on Learning Representations (ICLR)} (2023).

\bibitem{Perez2018FiLM}
E.~Perez, F.~Strub, H.~de~Vries, V.~Dumoulin, and A.~Courville.
FiLM: Visual Reasoning with a General Conditioning Layer.
In \emph{Proceedings of the AAAI Conference on Artificial Intelligence (AAAI)} (2018).

\bibitem{Zakharov1972JETP}
V.~E. Zakharov and A.~B. Shabat.
Exact theory of two-dimensional self-focusing and one-dimensional self-modulation of waves in nonlinear media.
\emph{Soviet Physics JETP} \textbf{34}(1), 62--69 (1972).

\bibitem{Sulem1999Springer}
C.~Sulem and P.-L. Sulem.
\emph{The Nonlinear Schr{\"o}dinger Equation: Self-Focusing and Wave Collapse}.
Springer Science \& Business Media (1999).

\bibitem{Weideman1986SIAM}
J.~A.~C. Weideman and B.~M. Herbst.
Split-step methods for the solution of the nonlinear Schr{\"o}dinger equation.
\emph{SIAM Journal on Numerical Analysis} \textbf{23}(3), 485--507 (1986).

\bibitem{Choi1998PRA}
S.~Choi, S.~A. Morgan, and K.~Burnett.
Phenomenological damping in trapped atomic Bose-Einstein condensates.
\emph{Physical Review A} \textbf{57}(5), 4057-4060 (1998).

\bibitem{Reeves2012PRA}
M.~T. Reeves, B.~P. Anderson, and A.~S. Bradley. 
Classical and quantum regimes of two-dimensional turbulence in trapped Bose-Einstein condensates.
\emph{Physical Review A} \textbf{86}, 053621 (2012).

\bibitem{GEK2026}
E.~Kiyani, A.~M. Deshpande, M.~Limaye, Z.~Gao, Z.~Zou,
S.~A. Pradeep, S.~Pilla, G.~Li, Z.~Li, and G.~E. Karniadakis,
Probabilistic predictions of process-induced deformation in carbon/epoxy composites using a Deep Operator Network,
\emph{Composites Part B: Engineering} \textbf{325}, 113952 (2026).


\end{thebibliography}
\end{document}